\documentclass[runningheads]{llncs}
\usepackage[T1]{fontenc}
\usepackage{graphicx}
\usepackage{booktabs}
\usepackage[misc]{ifsym}

\usepackage{amsmath,amssymb,amsfonts}
\usepackage{cite}
\usepackage{algorithmic}
\usepackage[linesnumbered,ruled,vlined]{algorithm2e}
\usepackage{graphicx}
\usepackage{textcomp}
\usepackage{xcolor}
\def\BibTeX{{\rm B\kern-.05em{\sc i\kern-.025em b}\kern-.08em
    T\kern-.1667em\lower.7ex\hbox{E}\kern-.125emX}}

\usepackage{hyperref} % url in references
\usepackage{cleveref} % Cref
\usepackage{graphicx} % figs
\usepackage{subcaption} % figs

\usepackage{thm-restate}

\newcommand*{\algo}{\texttt{P2NIA}\@\xspace}

\newcommand*{\algoBis}{\texttt{P2NIA}}

\usepackage{url}

\usepackage{breakurl}

\begin{document}

\title{\algo: Privacy-Preserving Non-Iterative Auditing}

\author{Jade Garcia Bourr\'ee\inst{1}\and
Hadrien Lautraite\inst{2} \and
S\'ebastien Gambs\inst{2} \and
Gilles Tredan\inst{3} \and
Erwan Le Merrer\inst{1} \and
Beno\^it Rottembourg\inst{4}}

\institute{Univ Rennes, Inria, CNRS, IRISA
\and
University du Québec à Montréal
\and
LAAS-CNRS
\and
Inria}

% Use \corr to indicate the corresponding author. Note the spacing around the \corr command. Only one author can be the corresponding author.
\authorrunning{J. Garcia Bourr\'ee et al.}

\maketitle

\begin{abstract} 
The emergence of AI legislation has increased the need to assess the ethical compliance of high-risk AI systems. 
Traditional auditing methods rely on platforms' application programming interfaces (APIs), where responses to queries are examined through the lens of fairness requirements.
However, such approaches put a significant burden on platforms, as they are forced to maintain APIs while ensuring privacy, facing the possibility of data leaks. 
This lack of proper collaboration between the two parties, in turn, causes a significant challenge to the auditor, who is subject to estimation bias as they are unaware of the data distribution of the platform.
To address these two issues, we present \algo, a novel auditing scheme that proposes a mutually beneficial collaboration for both the auditor and the platform. 
Extensive experiments demonstrate \algo's effectiveness in addressing both issues.
In summary, our work introduces a privacy-preserving and non-iterative audit scheme that enhances fairness assessments using synthetic or local data, avoiding the challenges associated with traditional API-based audits.

\end{abstract}

\section{Introduction}\label{s:introduction}
Algorithm auditing refers to the evaluation of algorithmic decision-making systems. 
More precisely, it aims at ensuring their privacy, transparency, fairness and compliance with ethical and legal standards~\cite{dunna2022paying, le2023algorithmic}. 
This field is very active, in reaction to algorithms becoming increasingly ubiquitous in our daily lives in critical areas such as finance, human resources, healthcare or justice~\cite{HBR2014social, chen2016empirical, financialtimes2019financial, silva2020facebook, guardian2023healthcare, Forbes2024other, BBC2024humanressources, CBS2024justice, ACC2024other}. 

The traditional way to assess if decision-making systems and models in production satisfy ethical standards is to audit them in the so-called \textit{black-box setting}, in which the auditor sends queries to a platform, receives answers and infers information on its behavior with respect to fairness, for instance.

\paragraph{Issues with black-box audits.}
Unfortunately, it has been shown that, in many real-world scenarios, black-box audits may not lead to accurate evaluation~\cite{barocas2023fairness,birhane2024ai, casper2024black}. The main reason is the non-collaboration of platforms that do not release information about their data distribution, kept hidden for privacy or trade secret motives.
Consider, for instance, the scenario of a bank who decides whom to lend to by using a model. The bank uses the model to predict if clients are likely to repay the loan based on their economic profile. An auditor wants to assess the fairness of such a model.
To realize this, requests composed of possible profiles are sent to the platform for assessment.
In this scarce data regime, audit results can lead to erroneous conclusions, possibly due to the auditor's measurement bias, whether deliberate or subconscious or to wrong assumptions regarding the audited model. 
For instance, if the auditor's requests do not align with the data distribution used to train the platform model, it will lead to biased conclusions.
In addition, many audit scenarios deal with sensitive data such as personal health records, income levels or demographic information like age, gender, and ethnicity~\cite{france2020article434_10}. 
Such data is highly critical and causes platforms to be reluctant to open and maintain APIs to expose them, as asked by recent legislation such as the AI act~\cite{IAAct}.

\paragraph{Contributions.} We propose a novel scheme to address these problems in which the platform and the auditor collaborate for the auditing.
This leads to a mutually beneficial situation in which 1) the auditor can perform an unbiased estimation of the property of interest (such as fairness in this paper), and 2) the platform does not need to maintain APIs while ensuring the privacy of its data. 
More precisely, we first demonstrate both theoretically and with an example that in a non-collaborative audit setup, where the auditor faces a black-box setting (\emph{i.e.}, without being provided the data distribution of the platform), an auditor obtains a biased estimate of the model under scrutiny. 
This motivates our scheme \algo (standing for Privacy-Preserving Non-Iterative Auditing), in which the platform collaborates with the auditor by releasing a synthetic dataset mimicking its behavior, allowing for unbiased audits while ensuring privacy. \\

In summary, our contributions are as follows:
\begin{itemize}
    \item We demonstrate and illustrate that an audit outcome can be biased in a black-box setup due to the \textit{population bias}.
    \item To circumvent this issue, we propose \algo, a novel collaborative auditing scheme benefiting both the platform and the auditor by enabling accurate audits while being privacy-preserving and non-iterative.
    \item We design several ways to implement \algo and experimentally evaluate their performance.
\end{itemize}

\paragraph{Outline.} First, we present the challenges associated with black-box auditing schemes in \Cref{s:setting} before introducing \algo as an alternative approach to address them in \Cref{s:P2NIA}. 
Then, we evaluate our proposal on two datasets and compare it to a standard audit scheme in \Cref{s:experiments}. 
Afterward, we review related works in \Cref{s:related work} before concluding in \Cref{s:conclusion}.

\section{Auditing Black-Box Models}
\label{s:setting}

After formalizing the audit objective in this section, we focus on the potential bias in black-box fairness estimation. 
In particular, studying a common hypothesis that an auditor needs to have knowledge of the platform distribution, we theoretically show that imperfect knowledge of this distribution can lead to a linear estimation bias on the statistical distance between the two distributions.
We also illustrate how such bias will likely appear in practice through a simulation.

\subsection{Auditing Fairness Setup}

\paragraph{Fairness Definitions.} The concept of fairness refers to the property that algorithms should not discriminate against people based on sensitive characteristics, such as ethnical origin or gender\cite{barocas2023fairness}.
Fairness can be defined in many ways, but we focus on \emph{group} fairness metrics, which compare the statistics of predictions between two groups, one of which is considered potentially discriminated and is called the protected group. 
In particular, we focus on \textit{demographic parity}, \textit{equality of opportunity} and \textit{equalized odds}, as these are arguably the most studied group fairness notions \cite{mehrabi2021survey}.

Formally, we audit model $m \in \mathcal{H}: \mathcal{X} \rightarrow \mathcal{Y}$, with $\mathcal{X}$ is its input space. 
Let $\mathcal{D}$ be the input distribution and $\mu_{\mathcal{D}}$ the target measure of the audit. 
Usually, this notation is shorthanded by $\mu$. 
However, this notation hides a strong prerequisite for measuring fairness, namely having access to the input distribution $\mathcal{D}$. 
In addition, group fairness requires the definition of a protected group, usually defined through a protected attribute $A\in\{0,1\}$. We denote $Y$ as the target variable and $\hat{Y} = m(X)$ as the prediction of model $m$ on input $X$.

The fairness metrics we study can be formally defined as follows~\cite{dwork2012fairness, hardt2016equality, besse2022survey}:\\
Demographic parity: {\small$\mu_{\mathcal{D}} = \left |P_{\mathcal{D}}[\hat{Y} = 1 | A = 1] - P_{\mathcal{D}}[\hat{Y} = 1 | A = 0]\right |$.}\\
Equalized odds: {\small $\mu_{\mathcal{D}} = \underset{y \in \{0,1\}}{max} \left| P_{\mathcal{D}}[\hat{Y}= 1 | Y = y, A = 0] - P_{\mathcal{D}}[\hat{Y} = 1 | Y = y, A = 1]\right|.$}\\
Equality of opportunity: {\small$\mu_{\mathcal{D}} =\left| P_{\mathcal{D}}[\hat{Y}= 1 | Y = 1, A = 0] - P_{\mathcal{D}}[\hat{Y} = 1 | Y = 1, A = 1]\right|. $}
 
In a nutshell, each of these metrics addresses different facets of potential bias and ensures that models perform equitably across groups. 
For instance, demographic parity aims at maintaining similar outcome distributions across groups, while equalized odds and equal opportunity assess fairness by comparing error rates with the latter, specifically focusing on true positive rates. Note that the last two fairness metrics require access to the true label $Y$ of samples, which is a strong assumption.

\paragraph{Black-Box Auditing Setting.} 
When assessing the fairness of a model, particularly in usual non-collaborative setups, auditors are unlikely to possess information about the distribution of training data.
Hence, they will have to rely on black-box auditing strategies, which generally entails crafting a specific finite set of queries $Q = \left\{ q_1, \dots, q_n \right\}$ in the model input space $\mathcal{X}$, and obtaining the corresponding honest\footnote{Previous works have shown that manipulative platforms may impede fairness assessment \cite{fukuchi2020faking}.} model inferred answers $m(Q) = \left\{m(q_1), \dots, m(q_n)\right\}$. 
The pair $(Q,m(Q))$ is then processed by the auditor to produce the fairness estimates. 
In the specific case of equality of opportunity and equalized odds,  the regulator typically requires domain expertise to access the true outcome $Y$ of their requests $Q$.

A crucial assumption for the auditing to work is that $Q \sim \mathcal{D}$, which is equivalent to the auditor having sampling access to the input distribution\footnote{In some settings, the auditor may want to deviate from $\mathcal{D}$ to optimize his sampling strategy and obtain more accurate estimates~\cite{de2024fairness}, but this nevertheless requires knowing $\mathcal{D}$.}.
Unfortunately, in a non-collaborative audit, the platform will likely keep this information confidential for privacy and trade secret reasons, as observed with recent legislation enforcing the opening of APIs \cite{IAAct}.

\subsection{Estimation Discrepancy Due to Population Bias}

When the auditor has no access to $\mathcal{D}$, they must use an approximation $\mathcal{D}'$ of $\mathcal{D}$ as a prerequisite to sample their queries. 
Unfortunately, the resulting fairness estimation will be impacted when $\mathcal{D}$ and $\mathcal{D}'$ differ significantly, as it will result in an inaccurate fairness estimation $\mu_{\mathcal{D}}(m)\neq \mu_{D'}(m)$.
Hereafter, we refer to this bias as a \emph{population bias}. This term was initially introduced to qualify the bias arising when a population $x\sim \mathcal{D}$ uses a classifier trained using samples drawn from a distribution $\mathcal{D}'\neq \mathcal{D}$~\cite{mehrabi2021survey}.
We use \textit{population bias} to qualify the bias due to the evaluation of models using a biased query set, \emph{i.e.} from a trained model on the platform using $x \sim \mathcal{D}$, then audited with a population $D'$.
While this form of bias has been known in other contexts, to the best of our knowledge it has not been studied in the field of model audits in the black-box setting. 
For instance,~\cite{casper2024black} acknowledges the possible presence of bias in the black-box setup but without identifying the precise bias cause.
To highlight the impact of such population bias in black-box audits, we first construct pathological examples demonstrating to which extent the estimation can be biased, 
before showing through simulation that population bias also has practical impacts, even in simpler scenarios.

\begin{restatable}[]{theorem}{unboundBias}\label{cor:unboundBias} \textbf{Population bias in black-box audit.}
Estimating the fairness metric (\emph{e.g.}, demographic parity, equality of opportunity or equalized odds) of a model $m$ using a distribution $\mathcal{D}'$ at a (total variation) statistical distance  $\alpha$ from the true distribution $\mathcal{D}$ can lead to an approximation error linear in $\alpha$. 
Formally: 
    $\exists \mathcal{D},\mathcal{D}',m$ s.t. $\delta(\mathcal{D},\mathcal{D}') = \sup_{F\in \mathcal{F}} \vert \mathcal{D}(F) - \mathcal{D}'(F) \vert = \alpha$ and $m\in \mathcal{H}:\mathcal{X} \mapsto \mathcal{Y}$ s.t. $\vert \mu_{\mathcal{D}}(m) - \mu_{\mathcal{D}'}(m) \vert = \alpha.$
\end{restatable}

The proof is given in \Cref{a:proof}.
Intuitively, this result states that if the auditor has a bias in their prior about the distribution at play at the platform (\emph{i.e.}, imperfect knowledge of the distribution), then the outcome is also a biased estimation. 
This is highly undesirable as it undermines the strength and legal impact of the audit.

\begin{figure}[!ht]
    \centering
    \includegraphics[width=\linewidth]{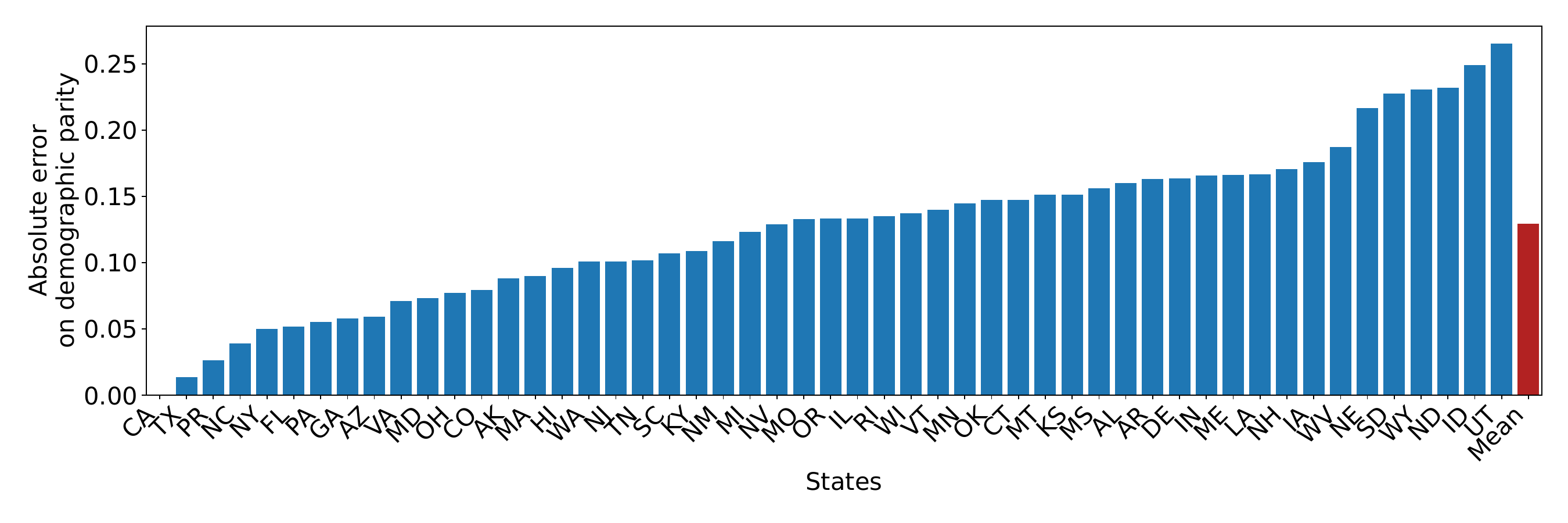}
    \caption{Observed population bias on the evaluation of demographic parity with protected feature ``sex'' on a model trained in state CA to predict whether an individual's income is above $\$50,000$ and evaluated in other states. 
    These results confirm the practical impact of \Cref{cor:unboundBias}.}
    \label{fig:FolktablesAcrossStates}
\end{figure}

While \Cref{cor:unboundBias} corresponds to a crafted extreme scenario, we now study the impact of population bias on real data. To this end, we use the standard Folktables dataset commonly used in fairness settings~\cite{ding2021retiring}.
Consider a platform that trained a model on data collected from a specific state (California) to predict whether an individual's income is above $\$50,000$~\cite{ding2021retiring}.
We consider a black-box auditor that has no access to this training distribution. Assume they use the distribution of another state among the remaining fifty in the dataset. 
One might expect that the auditor and model distributions are sufficiently similar to estimate the platform fairness accurately. 
However, as demonstrated by \Cref{fig:FolktablesAcrossStates}, this is not the case. 
Thus, no matter the state the auditor selects for reference, there will be a considerable population bias in the auditor's estimation. 
More concretely, on average, the absolute error in estimating the demographic parity with a different state is around $0.13$, which is high considering that the classical $80\%$ rule translates into a maximal tolerated value of $0.2$. Concretely, as the true demographic parity in this example is $0.01$, using the distribution of (NE,SD,WY,ND,ID,UT) would lead the auditor to a wrong conclusion about the model. 

\textbf{This population bias is thus a clear impediment for accurate auditing in non-collaborative black-box audits}, and is at stake even in such a simple and realistic scenario. 
To address this, in the following section, we propose \algo, a novel scheme permitting unbiased estimation while adding privacy protection with respect to the platform.

\section{Auditing through a Differentially-Private Dataset}
\label{s:P2NIA}

To tackle the aforementioned issues, we propose a collaborative scheme, \algo, which leverages privacy-preserving techniques for the benefit of the privacy protection of the platform while enabling accurate fairness auditing by the auditor.

\subsection{\algo: A Non-Iterative Auditing scheme}

We first present the \algo scheme before describing how it enables a fairness estimation that is 1) reliably performed and 2) without exposing the privacy of users (\emph{i.e.}, the released dataset is differentially-private).

\begin{figure}[!ht]
    \centering
    \includegraphics[width=\linewidth]{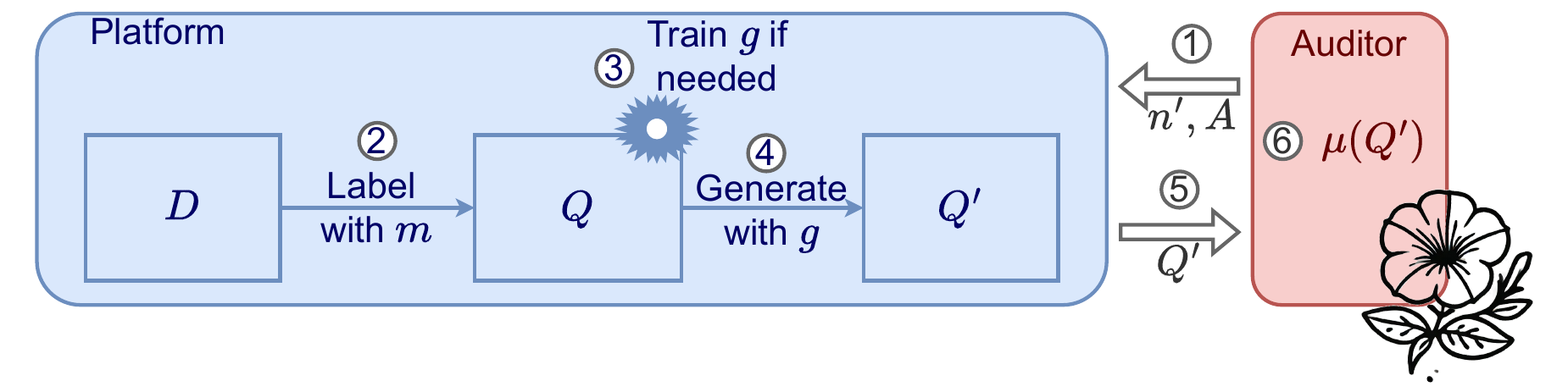}
    \caption{The \algo scheme steps by steps.}
    \label{fig:global-scheme}
\end{figure}

The Privacy-Preserving Non-Iterative Auditing scheme (\algo) works in six steps (\Cref{fig:global-scheme}).  
First, \algo starts with the auditor specifying to the platform the desired number of queries $n'$, along with $A$, the protected group of interest (\textbf{step 1}).

Upon receiving this request, the platform labels a part of its internal data (called \textit{audit dataset} in the following) using its proprietary model $m$. 
Model $m$ is applied to the dataset $D$ to generate a set of labeled queries $Q$, in which each query consists of a profile input $X$ from the dataset, the target output $Y$ and the corresponding predictions $\hat{Y}$ (\textbf{step 2}).

To protect sensitive information while still allowing fairness evaluation, the platform has two options in our framework. 
More precisely, either it anonymizes the dataset by using a local differential-private mechanism as described in \Cref{ss:DP-mechanisms}, or it trains a generative model $g$ build to produce synthetic data based on $Q$ to reflect the model's behavior without directly disclosing the training user data (\textbf{step 3}).
Afterward, the platform creates the audit dataset by generating or anonymizing $n'$ synthetic queries $Q'$ with $g$ (\textbf{step 4}).

Then, the platform releases $Q'$ to the auditor (\textbf{step 5}). 
At this stage, the auditor receives the auditing dataset and evaluates fairness using a predefined fairness metric $\mu (Q')$, applied to the protected group $A$ (\textbf{step 6}).\\

\par Observe that in our scheme, the auditor evaluates the model using $Q'$, without directly querying the platform model (step 6) but rather by declaring their number of requests and their group of interest (step 1). 
Thus, as the platform has generated $Q'$ from its audit dataset $D$ with a mechanism that ensures differential privacy, the personal data remains confidential while making it possible to audit the model (steps 2-5). 
As differential-privacy is immune to post-processing~\cite{dwork2014algorithmic} (i.e. it is impossible to compute a function of the output of the
private algorithm and make it less differentially private), the assessment of $\mu$ by the auditor does not reduce the privacy of the released data. For example, in the context of the income prediction model discussed in \Cref{s:introduction}, this would amount to accurately auditing the fairness of a platform based on synthetic demographic data without revealing the record of any user.

\subsection{Implementing \algo with Differential-private Mechanisms}\label{ss:DP-mechanisms}

The \algo scheme is generic with respect to particular privacy mechanisms implementations captured as $g$ in~\Cref{fig:global-scheme}. 
In this paper, we focus on differential privacy and explore two distinct mechanisms to achieve it: local differential privacy and synthetic data generation. 
We first present the differential privacy guarantees that we aim to achieve. 

\paragraph{Differential privacy.}
Differential privacy is a formal privacy model that protects individuals by bounding the impact any individual can have on the output of an algorithm\cite{dwork2014algorithmic}. 

\begin{definition}
    \textbf{Differential Privacy.~\cite{dwork2014algorithmic}} A randomized mechanism $\mathcal{M}$ satisfies $\epsilon$-differential privacy if for any individual $x\in \mathcal{X}$, any dataset $Q\subset \mathcal{X}$ and any subset of possible outputs $S$:
    $$
    P\left[\mathcal{M}(Q \setminus \{x\}) \in S\right] \leq exp(\epsilon) P\left[\mathcal{M}(Q) \in S\right]
    .$$
\end{definition}
The parameter $\epsilon$ is called the privacy budget and is considered public. 
The smaller $\epsilon$, the stronger the privacy guarantees. Typical values of $\epsilon$ could be $0.01$, $0.1$, or in some cases, $ln(2)$ or $ln(3)$~\cite{dwork2011firm}. 
However, in practice, $\epsilon$ can vary up to $10$~\cite{makhlouf2024systematic}, as, for example, Apple reports using values of $4$ or $8$~\cite{AppleDifferentialPrivacy}.

\paragraph{Data anonymization through local differential privacy.} 
One way to share data while respecting differential privacy is to add noise to the data. 
For instance, in the randomized response technique~\cite{warner1965randomized}, a respondent answers a sensitive binary question (\emph{e.g.}, ``What is your gender'') by tossing a biased coin in secret.
If the result is tail, the respondent answers honestly. Otherwise, the coin is tossed again, and the answer returned is ``Male'' if head and ``No'' if tail. 
This ensures that the auditor can reliably estimate the unknown proportions of sensitive attributes without knowing any individual's true answer with certainty. 
With Simple Random Sampling With Replacement (SRSWR~\cite{cochran1977sampling}), this generalized randomized response method provides an unbiased estimator of these proportions while also ensuring compliance with the $\epsilon$-differential privacy for any $\epsilon$ by adjusting the bias of the coin. 
The randomized response technique can be generalized to non-binary attributes (\emph{e.g.}, age or marital status) or to sets of attributes~\cite{chaudhuri2020randomized}, which is called Generalized Randomized Response (GRR). 
In addition, it can be combined with pre-processing techniques~\cite{wang2023privacy} (instead of SRSWR) for better privacy guarantee, and recent advances have further improved the performance of GRR~\cite{erlingsson2014rappor,  bassily2015local, wang2017locally, ye2018optimal, domingo2020multi}.

\par - \textit{\algo implementation:} Since our study aims to highlight how \algo can be applied rather than providing an exhaustive comparison of private data methods, we proceed with the basic version of GRR: each colmun of the dataset (features, target and prediction) is independently flipped with a probability $p$ selected to reach the desired privacy guarantee $\epsilon$: $p=e^\epsilon/ (e^\epsilon + k -1)$ where $k$ is the number of values the feature can have.

\par - \textit{\algo Fairness Estimation Reliability:} As $\hat{Y}$ is computed by the platform knowing $D$, the resulting fairness estimator is only biased by the noise added by GRR. The auditor knows $\epsilon$ and hence $p$, so they can easily unbias the anonymized results to recover the correct estimator value \cite{chaudhuri2020randomized, domingo2020multi}.

\paragraph{Mechanisms for differentially-private synthetic data generation.} 
Instead of adding noise to existing data, another approach consists of generating synthetic data that follows the original distribution while protecting the privacy of individual records for the original dataset. 
Many differentially-private data generation approaches have been proposed in the literature to achieve this objective; the interested reader can refer to\cite{tao2021benchmarking} or~\cite{figueira2022survey} for recent surveys on the subject. In particular, some synthesis methods offer the possibility to specify which relationship between variables should be maintained, aligning with our goal of reliable model evaluation.
Specifically, we use MST\cite{mckenna2021winning} and AIM \cite{mckenna2022aim}, which are two methods operating under the \textit{select-measure-generate} framework. In a nutshell, both methods work by first selecting a set of low-dimensional marginal queries (\emph{i.e.}, statistics on small subsets of attributes). Afterward, each selected marginal is measured under differential privacy by adding noise to ensure privacy. 
This yields a collection of noisy marginal results. Finally, a graphical model called Private-PGM \cite{mckenna2019graphical} is constructed from those noisy marginals to infer a consistent high-dimensional data distribution to generate new synthetic data points. 
MST and AIM differ in the way marginals are selected. 
More precisely, MST performs an initial noisy measurement of marginals using a small portion of the privacy budget and then selects which one should be measured more precisely. 
In contrast, AIM adaptively chooses new queries in iterative rounds, using feedback from previous measurements to guide where the privacy budget should be spent. 
This adaptive selection can improve accuracy at the expense of increased computational cost. 

\par - \textit{\algo implementation:} 
We explicitly maintain the relationship between protected variable $A$, target variable $Y$ and predictions $\hat{Y}$ from the audited model (\textit{i.e.} the marginals $(A, Y), (A, \hat{Y}), (Y, \hat{Y})$ and $(A, Y, \hat{Y})$) in addition to $12$ 2-way marginals chosen randomly to mimic the dataset. 
The relationship between the required privacy level $\epsilon$ and the noise introduced in the marginals differs between MST and AIM, which are detailed in the respective original papers.

\par - \textit{\algo Fairness Estimation Reliability:} As we explicitly conserve all marginals involved in the definition of all three fairness measures, the resulting value is unbiased.

Hence, we consider three distinct methods to implement \algo: GRR, MST and AIM. While the choice of GRR is driven by its appealing simplicity, MST and AIM are recent synthetic data generation methods. With each of these methods \textbf{\algo, the fairness estimation is reliably performed and respects differential privacy on user's data.} While each method guarantees the obtention of an unbiased fairness estimator (for an audit set $Q$ of infinite size), as we will see in the next section, each method has a different impact on the convergence of fairness estimators.

\section{Experimental Evaluation}
\label{s:exp}
\label{s:experiments}

While involving anonymized or synthetic data in audits has theoretical advantages compared to black-box audits, this section demonstrates the effectiveness of \algo in practice. To assess the performance of the proposed approach, we conducted experiments on two datasets commonly used in audit studies while varying the function $g$ within \algo. 
The results demonstrate that our proposed scheme using anonymized or synthetic data achieves significantly higher accuracy than black-box audits.

We first present our experimental setup before evaluating the advantage of using anonymized or synthetic data.

\subsection{Experimental Setting}

\paragraph{Datasets and models.} 
Experiments are performed on two standard benchmark datasets from the fair and private machine learning literature: Adult~\cite{adult_2} and Folktables~\cite{ding2021retiring} (the full
description is deferred to \Cref{a:data}). 

To solve the tasks outlined above, we employ random forests as implemented in the scikit-learn Python Library with its \textbf{RandomForestClassifier} object. 
Hyper-parameters were selected using a grid search to minimize the model’s loss on each dataset. 
Further experiments using gradient boosting algorithm (\textbf{XGBClassifier}) are provided in \Cref{a:exp}.

\paragraph{Experimental parameters.} All datasets were split into two parts: $80\%$ of the dataset as the training set for the model $m$ while the remaining $20\%$ is used as a test set. 
Only the test set is used as input of the function $g$ (step 4 in~\Cref{fig:global-scheme}).
This process is repeated ten times with the same random seeds for the train/test split and the same model, with the results averaged across these ten iterations. 
All experiments are conducted on a computing cluster of homogeneous nodes powered by Intel Xeon E5-2660 v2 processors.

\paragraph{Reference.} 
Since the objective is to evaluate the performance of \algo in the audit of fairness, we establish a reference by defining the true fairness value of the model used by the platform. 
Specifically, this reference corresponds to the fairness metrics computed on the test set of the datasets once they have been labeled by the model in question. This is the target value that the auditor wants to estimate.

\paragraph{Baseline.} 
We compare our approach with the black-box scenario in the following manner. 
The auditor is assumed to have knowledge of the set of possible values for each attribute ($\mathcal{X}$), but not to the overall distribution of the data ($\mathcal{D}$). 
The auditor draws each attribute uniformly from its possible values to generate queries. 
The same approach is also applied to the outputs $Y$ when auditing fairness metrics such as Equalized Odds or Equality of Opportunity.\\

The performance of \algo is compared to the black-box audit on the two binary classification tasks from Adult and Folktables. 
To realize this, we instantiate \algo using locally anonymized data or synthetic data generated by the methods GRR~\cite{chaudhuri2020randomized}, MST~\cite{mckenna2021winning} or AIM~\cite{mckenna2022aim} described in \Cref{ss:DP-mechanisms}. 
More precisely, we used publicly available implementations of those methods~\cite{PrivatePGM_2021}.

\subsection{Impact of Sample Size on Demographic Parity}

We study the practical effectiveness of \algo by evaluating its performance depending on the number of samples to assess the demographic parity. 
The differential privacy parameter $\epsilon$ is arbitrarily set to $10$.

\begin{figure}
\centering
\begin{subfigure}{.5\textwidth}
  \centering
  \includegraphics[width=\linewidth]{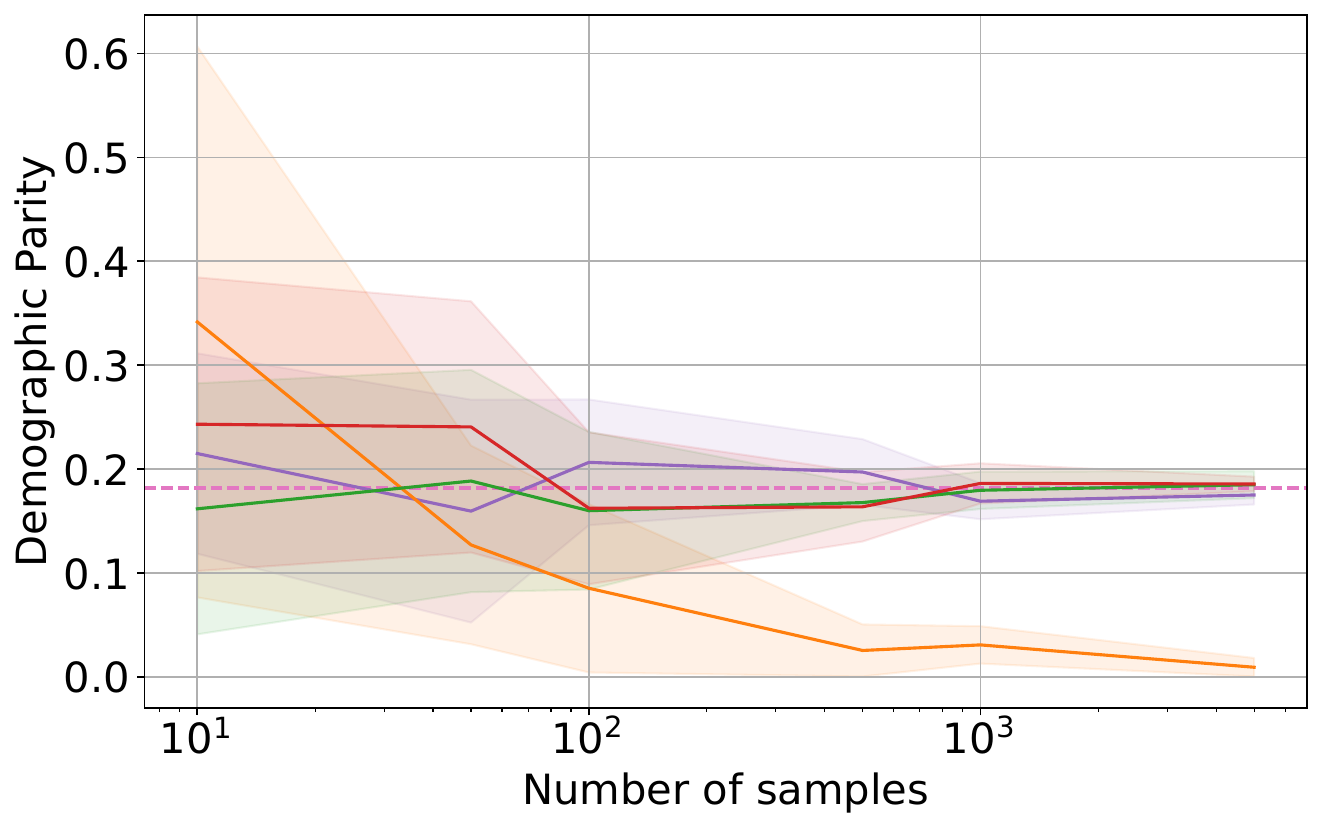}
  \caption{Adult}
\end{subfigure}%
\begin{subfigure}{.5\textwidth}
  \centering
  \includegraphics[width=\linewidth]{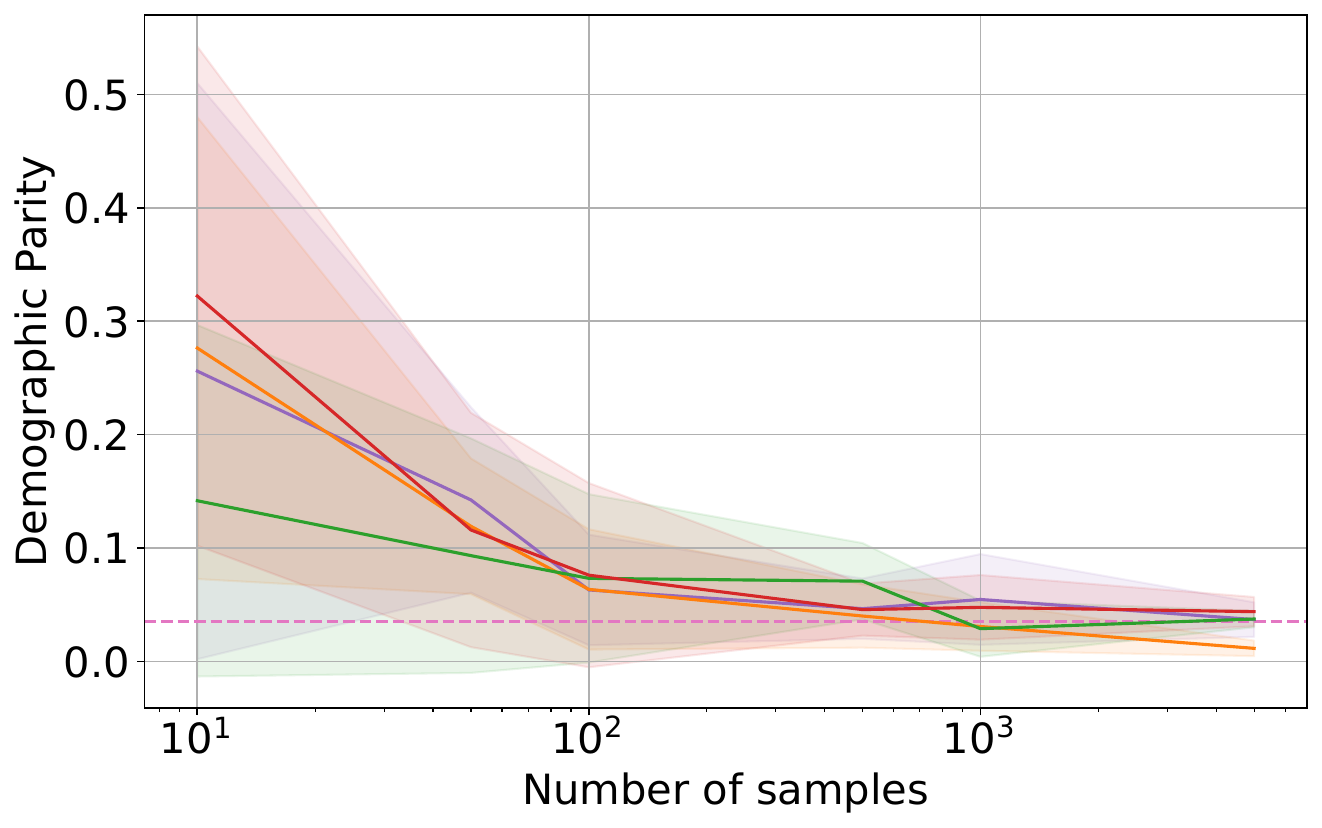}
  \caption{Folktables}
\end{subfigure}

\begin{subfigure}{\textwidth}
  \centering
  \includegraphics[width=.9\linewidth]{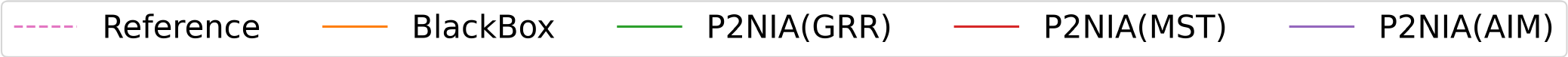}
\end{subfigure}%

\caption{Demographic parity depending on the number of samples.}
\label{fig:DePa-n}
\end{figure}

Analyzing the demographic parity depending on the number of samples (\Cref{fig:DePa-n}) shows that the convergence rate of the estimator remains largely unaffected by the method used within \algo. 
This indicates that the selection of the method within the algorithm does not significantly affect the rate at which the estimator converges.

It is noteworthy that while increasing the number of samples enables a more accurate estimation, \textbf{\algo achieves high precision even with a modest sample size}. 
This contrasts with the black-box approach, in which convergence is slower and tends towards an incorrect value, resulting in unreliable fairness assessments. 
This outcome highlights a significant constraint of the black-box approach, namely its reliance on a dataset of significant size to achieve even an approximate solution, which is inherently biased.

\subsection{Impact of Differential Privacy on Demographic Parity}
\begin{figure}
\centering
\begin{subfigure}{.5\textwidth}
  \centering
  \includegraphics[width=\linewidth]{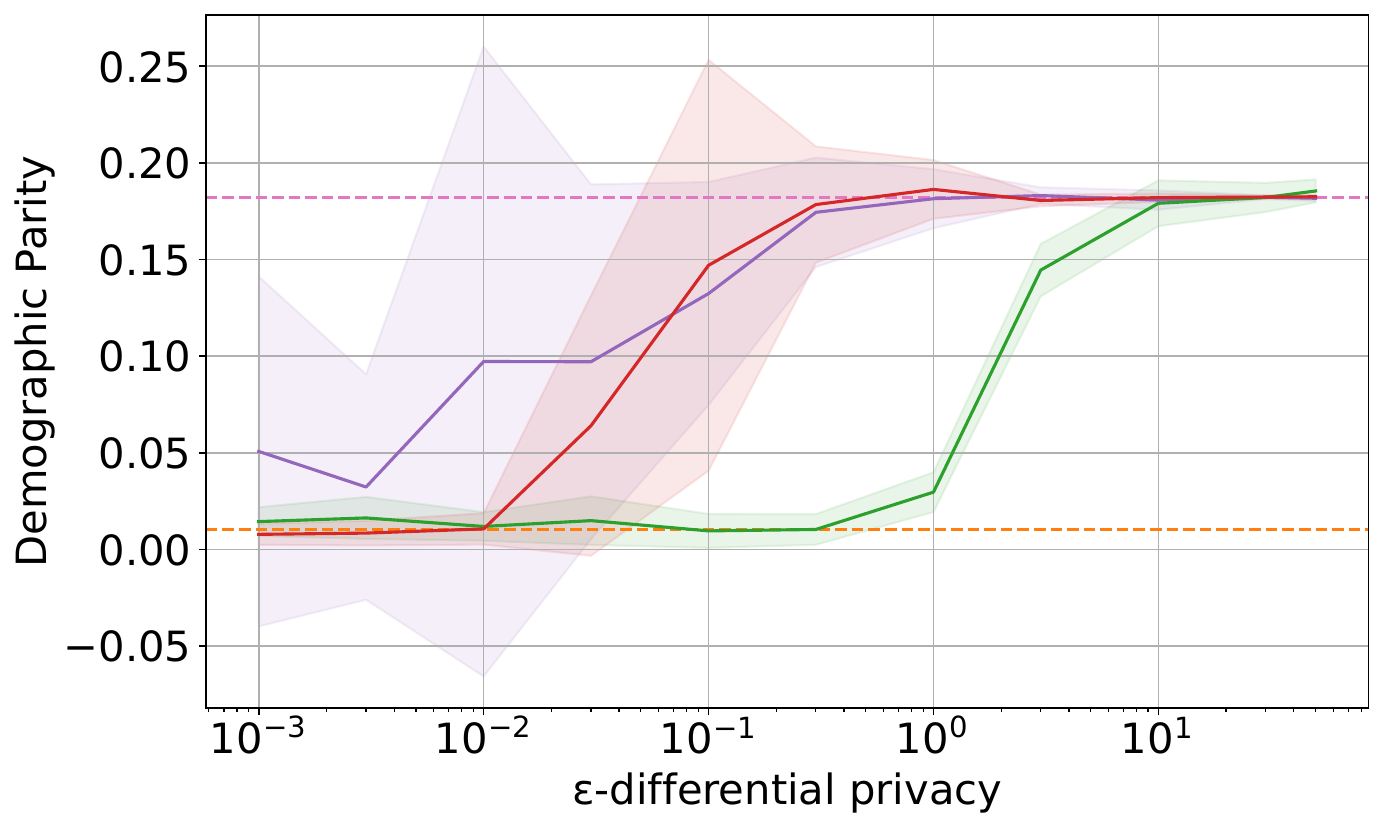}
  \caption{Adult}
\end{subfigure}%
\begin{subfigure}{.5\textwidth}
  \centering
  \includegraphics[width=\linewidth]{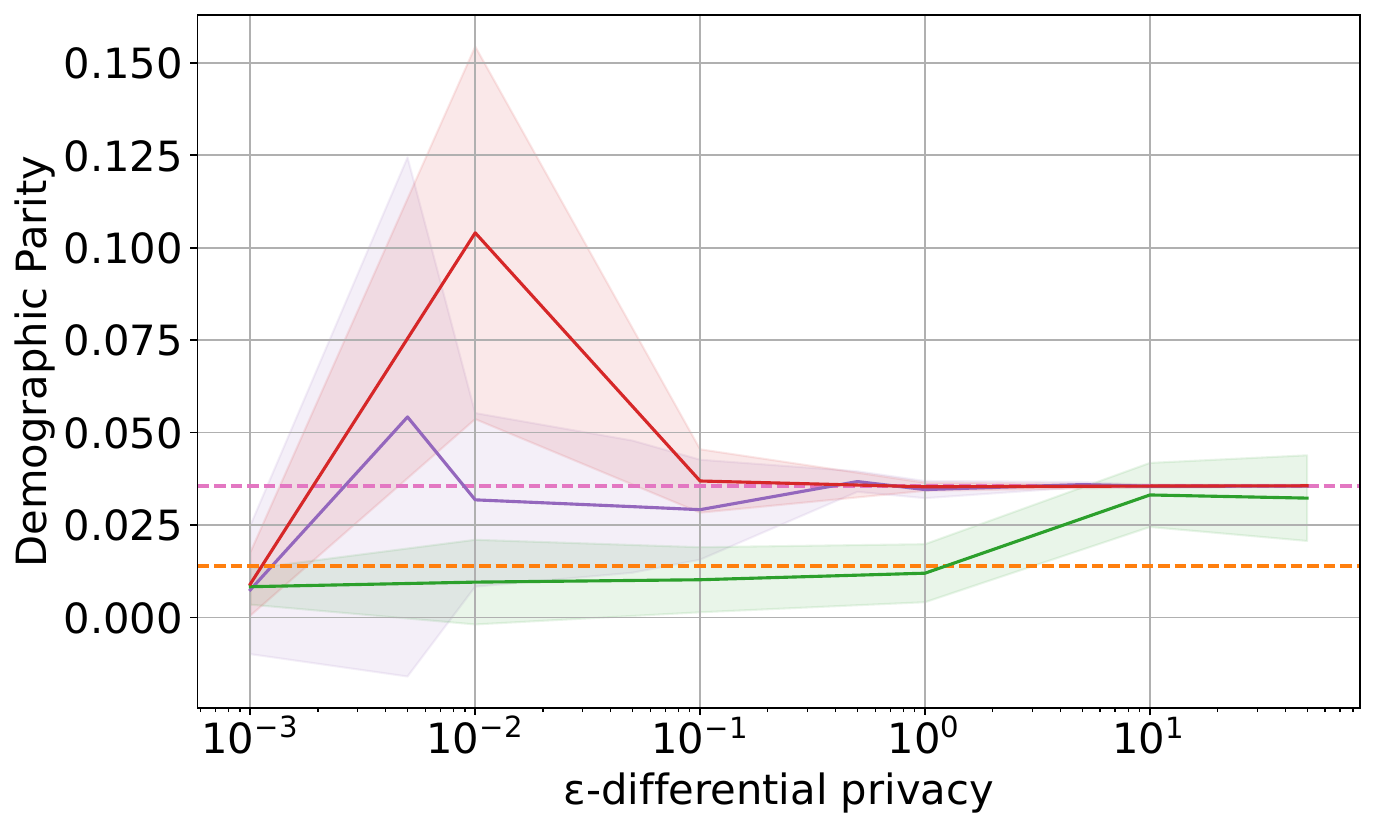}
  \caption{Folktables}
\end{subfigure}%

\begin{subfigure}{\textwidth}
  \centering
  \includegraphics[width=.9\linewidth]{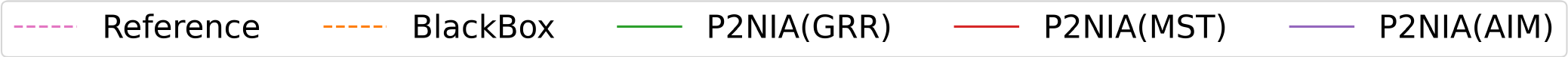}
\end{subfigure}%
\caption{Demographic parity depending on the $\epsilon$-differential privacy in \algo.}
\label{fig:DePa-DiPr}
\end{figure}

We have explored the average error on demographic parity (on $5,000$ samples) for different levels of differential privacy in \Cref{fig:DePa-DiPr}.

Regardless of the value of $\epsilon$, the black-box audit consistently produces the worst results for the Adult dataset, highlighting its inefficiency. 
Specifically, when the value of $\epsilon$ is large, the estimated demographic parity with \algo exhibits high accuracy, but with a low privacy level. 
Conversely, for small values of $\epsilon$, the accuracy of the demographic parity estimator decreases, but the privacy protection increases. 

On the Folktables dataset, \algo exhibits similar asymptotic behavior to that observed on the Adult dataset, reinforcing the generalizability of its performance. However, a key difference emerges in the stability of intermediate values of $\epsilon$, for \algo(MST) and \algo(AIM). In these cases, the lack of stability can be attributed to the complexity introduced by the noisy marginals, affecting the consistency of the fairness estimates. In contrast, \algo(GRR) demonstrates a more controlled and predictable behavior despite converging more slowly.

These findings reveal the existing trade-off between differential privacy and estimator accuracy on demographic parity. 
The optimal balance between these two metrics is likely to depend on the characteristics of the dataset and the generation method employed with realistic $\epsilon$ \Cref{ss:DP-mechanisms}.
Notably, \textbf{\algo enables effective platform audit of demographic parity}, a capability that the black-box approach lacks due to its inherent bias.

\subsection{Head-to-Head Comparison of \algo on Other Fairness Metrics}

In this section, we show how \algo can be used to audit other fairness metrics, such as equalized odds and equality of opportunity. 
To realize this, we set the differential privacy parameter $\epsilon$ to $10$. 
Additional experiments for $\epsilon = 1$ can be found in \Cref{a:exp}.  

\begin{figure}
\centering
\begin{subfigure}{.5\textwidth}
  \centering
  \includegraphics[width=\linewidth]{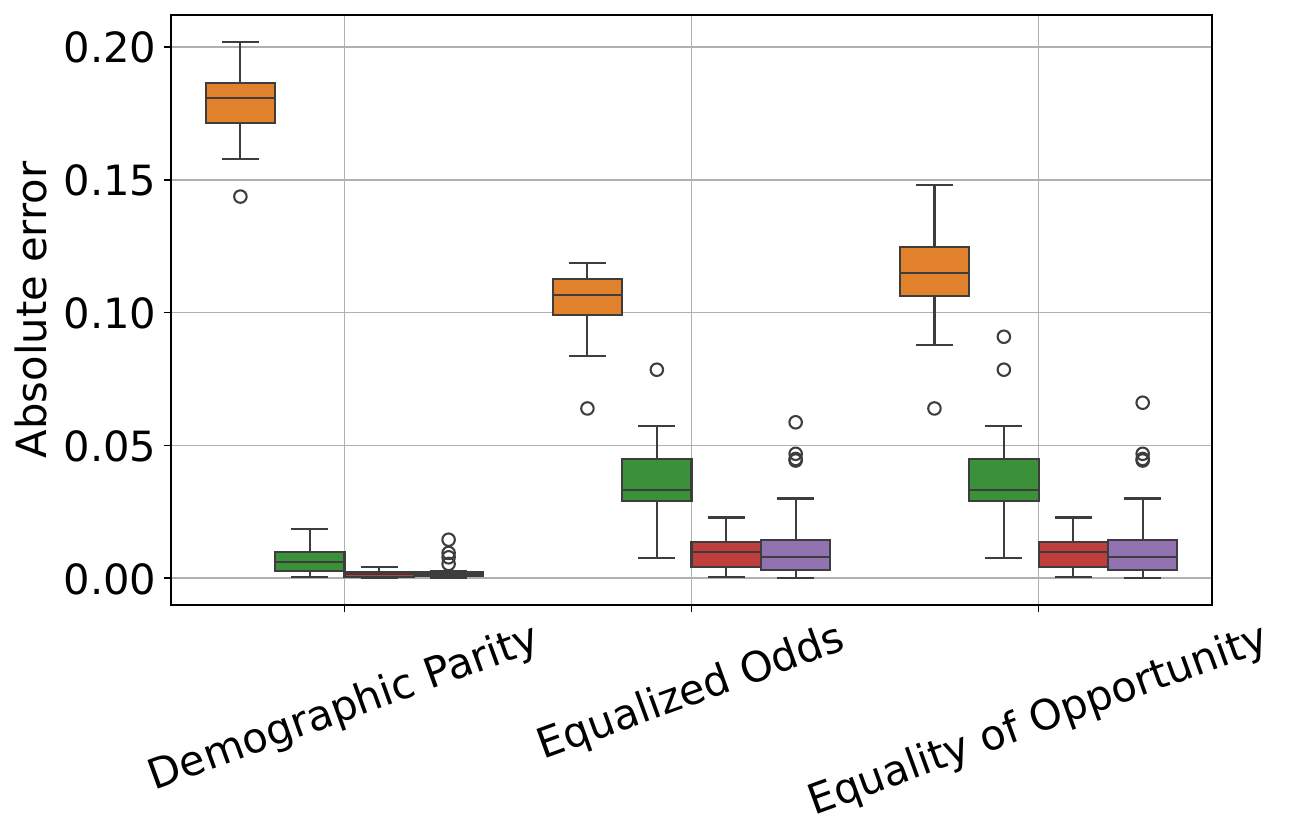}
  \caption{Adult}
\end{subfigure}%
\begin{subfigure}{.5\textwidth}
  \centering
  \includegraphics[width=\linewidth]{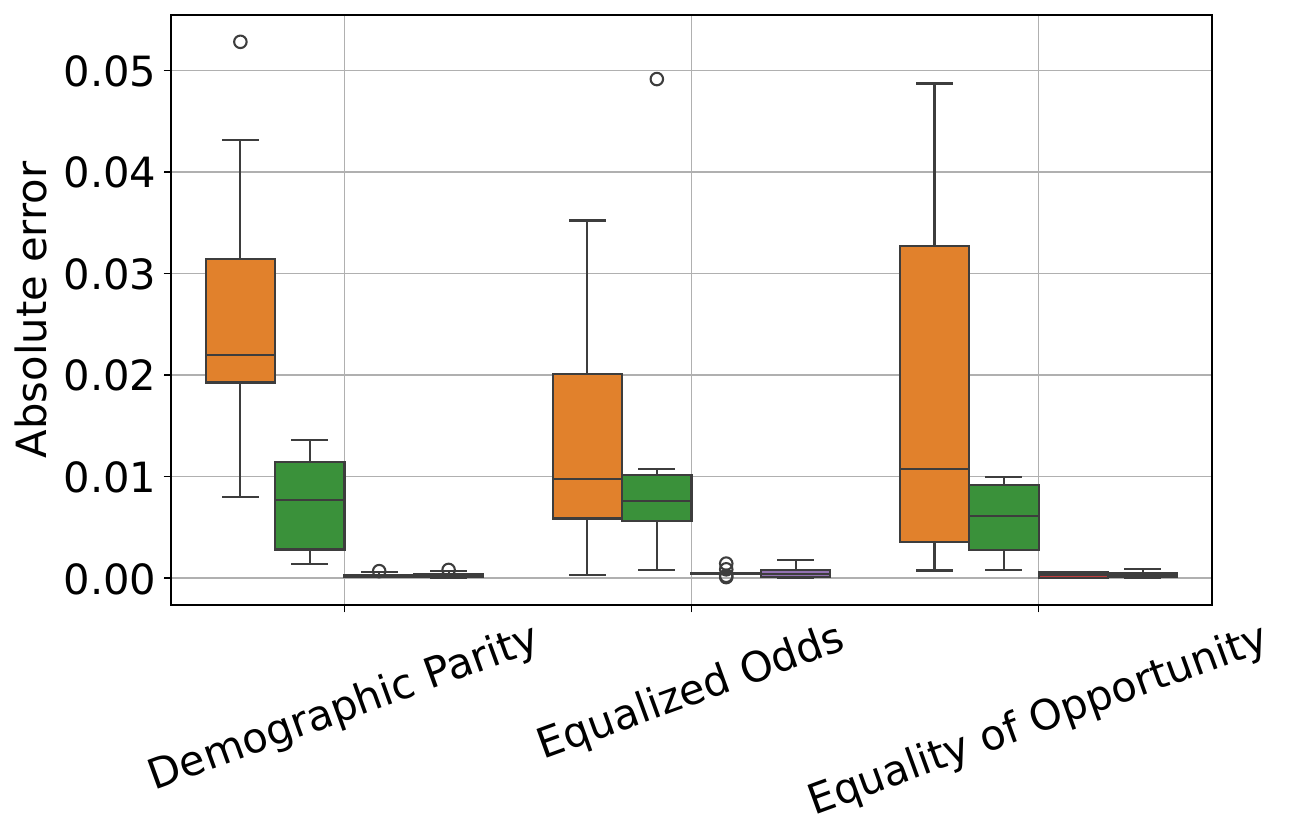}
  \caption{Folktables}
\end{subfigure}

\begin{subfigure}{\textwidth}
  \centering
  \includegraphics[width=.9\linewidth]{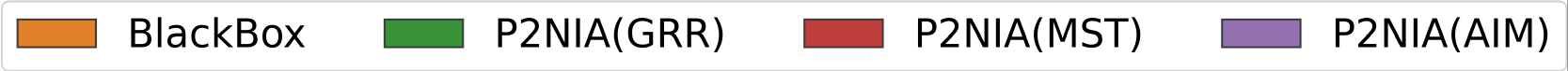}
\end{subfigure}%
\caption{Absolute difference compared to the reference for the three standard fairness metrics with $\epsilon = 10$, averaged on the two models.}
\label{fig:boxplot}
\end{figure}

For a constant value of $\epsilon = 10$, \Cref{fig:boxplot} demonstrates that the earlier results extend beyond demographic parity to other fairness metrics, such as equalized odds and equality of opportunity. 
Specifically, our scheme with locally anonymized data (\emph{i.e.}, $\algoBis(GRR)$) performs comparably to, or even better than, black-box auditing. 
Consequently, by releasing locally anonymized data, the platform makes it possible to reliably estimate fairness. 
Furthermore, our scheme instantiated with synthetic data (\emph{i.e.}, $\algoBis(MST)$ and $\algoBis(AIM)$) also results in a significant error reduction. 
These findings highlight the usefulness of our approach as a fairness auditing tool, demonstrating \textbf{the ability of \algo to evaluate not only Demographic Parity but also Equalized Odds and Equality of Opportunity effectively}.

\section{Related Dwork}
\label{s:related work}
Fairness auditing, particularly in black-box settings, has been extensively explored~\cite{sandvig2014auditing, chen2016empirical, silva2020facebook, bandy2021problematic, dunna2022paying}. These approaches rely on analyzing the outputs of the model based on a set of queries designed to assess fairness. For black-box auditing, the fundamental lever the auditor relies on is their choice of queries. 
This could involve i.i.d. sampling \cite{feldman2015certifying}, stratified sampling \cite{taskesen2021statistical} or more complex setups \cite{panigutti2021fairlens}. 
However, all these methods rely on the central assumption that the auditor knows the model distribution in advance, which is unrealistic due to the knowledge asymmetry between non collaborative platforms and auditors.

Thus, there has some been recent work~\cite{birhane2024ai, casper2024black}, arguing for more transparent ways to evaluate the assessment of algorithms with respect to ethical standards as audits are imprecise due to the lack of comprehension of these algorithms. 

Alternative approaches for fairness auditing include, for instance, the white-box setting in which the model is provided to the auditor~\cite{barocas2023fairness, casper2024black, casper2024black} or fairness certification~\cite{shamsabadi2022confidential, kang2022certifying, duddu2023attesting}. 

As fairness auditing often requires querying the model on user data, privacy concerns have also emerged as a fundamental issue. 
Dwork et al. \cite{dwork2012fairness} has already identified and prove the link between individual fairness and local differential privacy. It has been studied in the context of fair learning~\cite{makhlouf2024systematic} (\textit{e.g.} before model being in production) but not for audit purpose. 
However, \cite{kilbertus2018blind, pentyala2022privfair} introduced differential privacy techniques to mitigate the risk of information leakage during fairness audits. 
Other methods, such as secure multi-party computation~\cite{toreini2023verifiable} rely on complex cryptographic techniques to ensure the privacy of queries. 
A key limitation of these previous works is that they focus on protecting the auditor’s queries rather than the data of the entity being audited. 

Our approach, \algo, aims to bridge this gap by allowing effective fairness auditing without compromising user privacy.

\section{Conclusion}
\label{s:conclusion}

To summarize, in this paper, we have first underlined an important problem for black-box audits in real scenarios: population bias due to the lack of a perfect prior for the auditor on the data distribution of the platform. 
This arises mainly due to the lack of collaboration between the platform and the auditor. 
We then proposed a novel audit scheme, named \algo, in which both parties collaborate out of mutual benefit, namely accurate estimation for the auditor and data privacy preservation plus ease of operation for the platform as the scheme is non-iterative. 
We empirically show on standard datasets that our scheme operates as intended for three group fairness metrics, with accurate assessment and controllable privacy guarantees.

Future work includes the study of other forms of collaborations driven by different benefits at each party. 
For instance, it might be beneficial for the platform to operate in a ``grey-box'' setting, in which it will reveal some information about the internal workings of its model, resulting in a more lightweight querying procedure by an auditor (\emph{e.g.}, with respect to the number of queries required). 

\bibliographystyle{splncs04}
\bibliography{references}

\clearpage

\appendix

\section{Black-box Setting Bias}
\label{a:proof}

\unboundBias*

\textbf{Proof.} First, we deal with demographic parity.
We proceed by constructing the situation considered.
Let $X = a\in\{0,1\} \times b\in \{0,1\} \times c\in [0,1]$ be the input space and $Y=\{0,1\}$ the binary output space. 
In the following, $a$ will play the role of the protected attribute, $b$ will enable the tuning of the statistical distance between $D$ and $D'$, and $c$ represents additional data used by the model to construct its decision. 
We now describe two distributions $D,D'$ on $X$:
\newcommand{\ind}{\perp\!\!\!\perp} 
\begin{itemize}
    \item for both $D$ and $D'$ we have: $c\ind (a,b), c\sim U(0,1)$ and $a\ind b$.
    \item $\mathbb{E}_{D} (a=1) = \mathbb{E}_{D'} (a=1) = 1/2$
    \item $\mathbb{E}_{D} (b=1) = 0$ and $\mathbb{E}_{D'} (b=1) = \alpha$ 
\end{itemize}
As a table:
\begin{table}[h!] \centering
\begin{tabular}{l|llll}
\textbf{$x \sim$} & $a=1,b=1$  & $a=0,b=1$  & $a=1,b=0$      & $a=0,b=0$      \\ \hline
$D$               & 0          & 0          & $1/2$          & $1/2$          \\
$D'$              & $\alpha/2$ & $\alpha/2$ & $(1-\alpha)/2$ & $(1-\alpha)/2$
\end{tabular}
\end{table}

Clearly, with $\mathcal{F}$ being a sigma-algebra of $X$, we have $$\delta(D,D') = sup_{F\in \mathcal{F}}|P_D(F)-P_{D'}(F)| = P_{D'}(b=1)-P_{D}(b=1) = \alpha.$$ 
In other words, the total variation distance between $D$ and $D'$ is $\alpha$. 

The model $M:X\mapsto Y$ is defined as follows: $M(x)=(1-b).(c>1/2) + b.a$. 
The rationale here is that $b$ drives two distinct behaviors of $M$, only one of which is unfair. 
The demographic parity can be expressed as given $D$:
$\mu_{x\sim \mathcal{D}}(M) = |P(Y|a)-P(Y|\bar{a})| = P(c>1/2) - P(c\leq 1/2) =0$. 
Conversely, given $D'$, using the law of total probability on the partitions induced by $b$: $\mu_{x\sim \mathcal{D}'}= (P(Y|a,b) - P(Y|\bar{a},b))P(b)  + (P(Y|a,\bar{b}) - P(Y|\bar{a},\bar{b}))(1-P(b)) = (1-0)\alpha +0 = \alpha$. 
Hence, $|\mu_{x\sim \mathcal{D}} (M) - \mu_{x\sim \mathcal{D}'}(M)| = \alpha$.

Second, we can generalize the previous proof to equality of opportunity and equalized odds by assessing that for all input, $Y = 1$. In such case, for all binary $a, y$, $P_{\mathcal{D}}[\hat{Y}= 1 | Y = y, A = a] = P_{\mathcal{D}}[\hat{Y}= 1 | Y = 1, A = a] = P_{\mathcal{D}}[\hat{Y}= 1 | A = a]$ and the three metrics are equals.
\hfill$\square$

\section{Descriptions of the data and models}\label{a:data}
The \textbf{Adult} dataset~\cite{adult_2} is commonly used for tasks related to classification and fairness in machine learning. The dataset contains demographic information from the 1994 U.S. Census with respect to 32.561 individuals, and the task is to predict whether an individual's income exceeds \$50,000 per year. The dataset contains $12$ attributes: age, employment status, education level, marital status, occupation status, relationship, ethnical origin, gender, capital gain, capital loss, working hours and native country. We consider the binary attribute ``gender'' to be the protected one in this dataset. 

The \textbf{Folktables} dataset~\cite{ding2021retiring} is derived from the U.S. Census data and is designed to evaluate the fairness of machine learning models. We deal with the task \textit{ACSIncome} that consists in predicting whether U.S. working adults' yearly income is above $\$50,000$. The dataset contains 378.816 samples and includes various demographic attributes such as age, worker class, educational attainment, marital status, occupation, place of birth, relationship, working hours, gender and ethnical origin. For a complete description of these attributes, refer to the Folktables documentation\cite{ding2021retiring}.

The \textbf{Random Forest Classifier} is configured with $30$ decision trees, a maximum of $11$ features considered for each split, and a maximum tree depth of $15$. This setup balances performance and computational efficiency while preventing excessive over-fitting.

For the \textbf{XGBoost Classifier}, we use $500$ boosting rounds with a learning rate of $0.05$. The model is optimized using the logarithmic loss function, and early stopping is applied after $5$ rounds to prevent over-fitting.

Both architectures are used with both datasets with the same hyper-parameters but independently trained for each dataset.

\section{Additional experiments}\label{a:exp}

The following section contains additional figures not referenced in the body of the paper. All analyses are valid on these new figures. They contain variants of Figures \ref{fig:boxplot}, \ref{fig:DePa-n} and \ref{fig:DePa-DiPr} for the XGB model on both datasets and with $\epsilon = 1$ or $10$.

\begin{figure}
\centering
\begin{subfigure}{.5\textwidth}
  \centering
  \includegraphics[width=\linewidth]{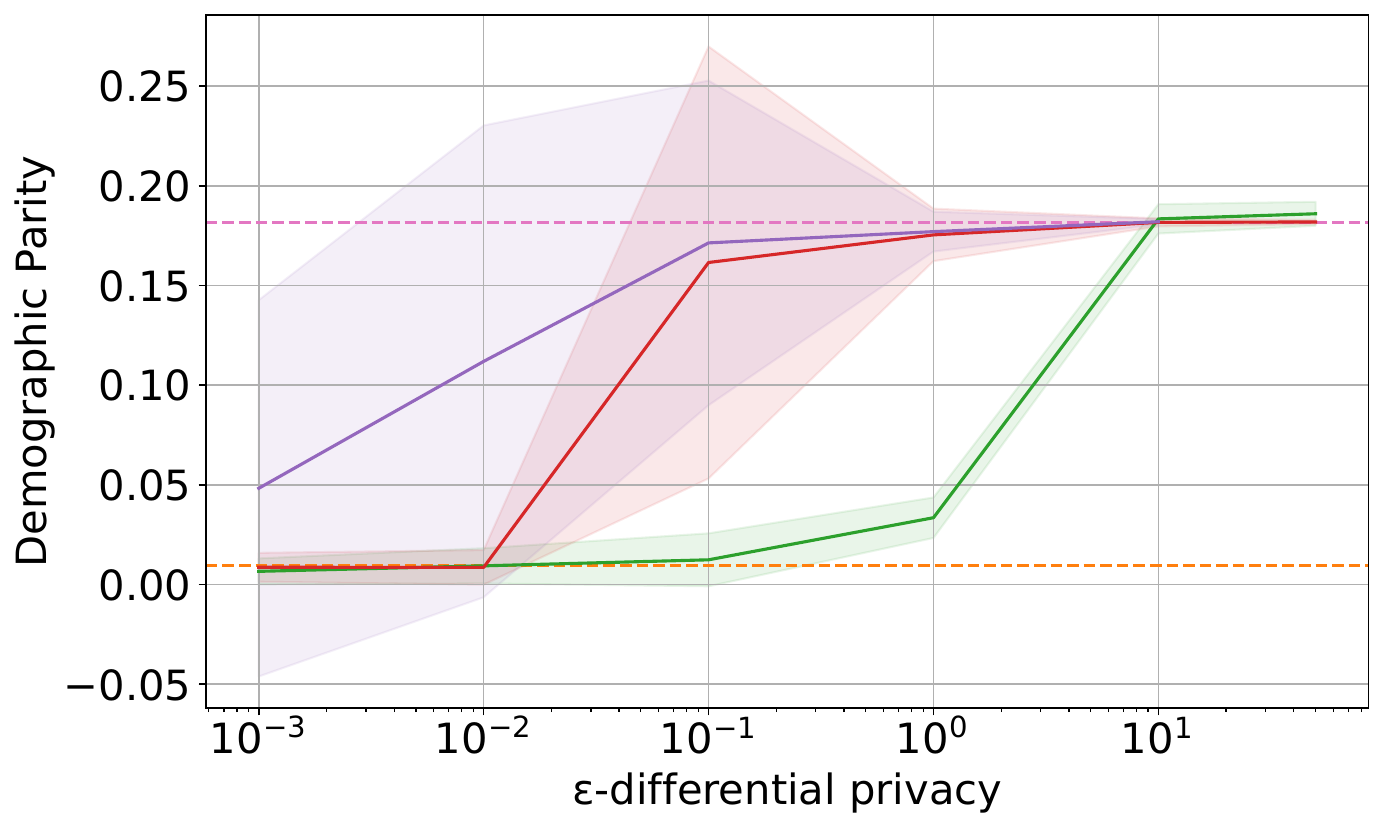}
  \caption{Adult}
\end{subfigure}%
\begin{subfigure}{.5\textwidth}
  \centering
  \includegraphics[width=\linewidth]{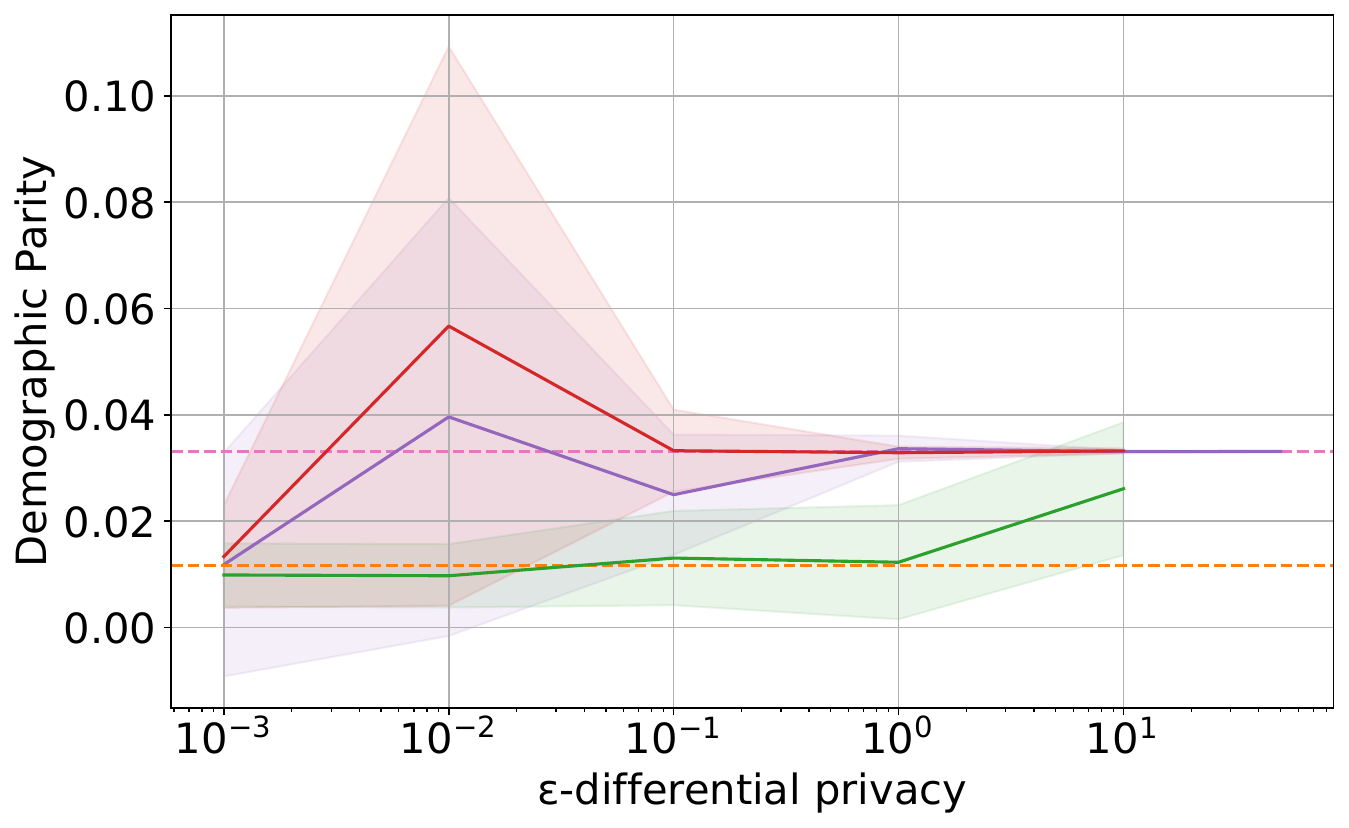}
  \caption{Folktables}
\end{subfigure}%

\begin{subfigure}{\textwidth}
  \centering
  \includegraphics[width=.9\linewidth]{pics/legende_epsilon.pdf}
\end{subfigure}%
\caption{Demographic parity depending on the $\epsilon$-differential privacy with XGB.}
\end{figure}

\begin{figure}
\centering
\begin{subfigure}{.5\textwidth}
  \centering
  \includegraphics[width=\linewidth]{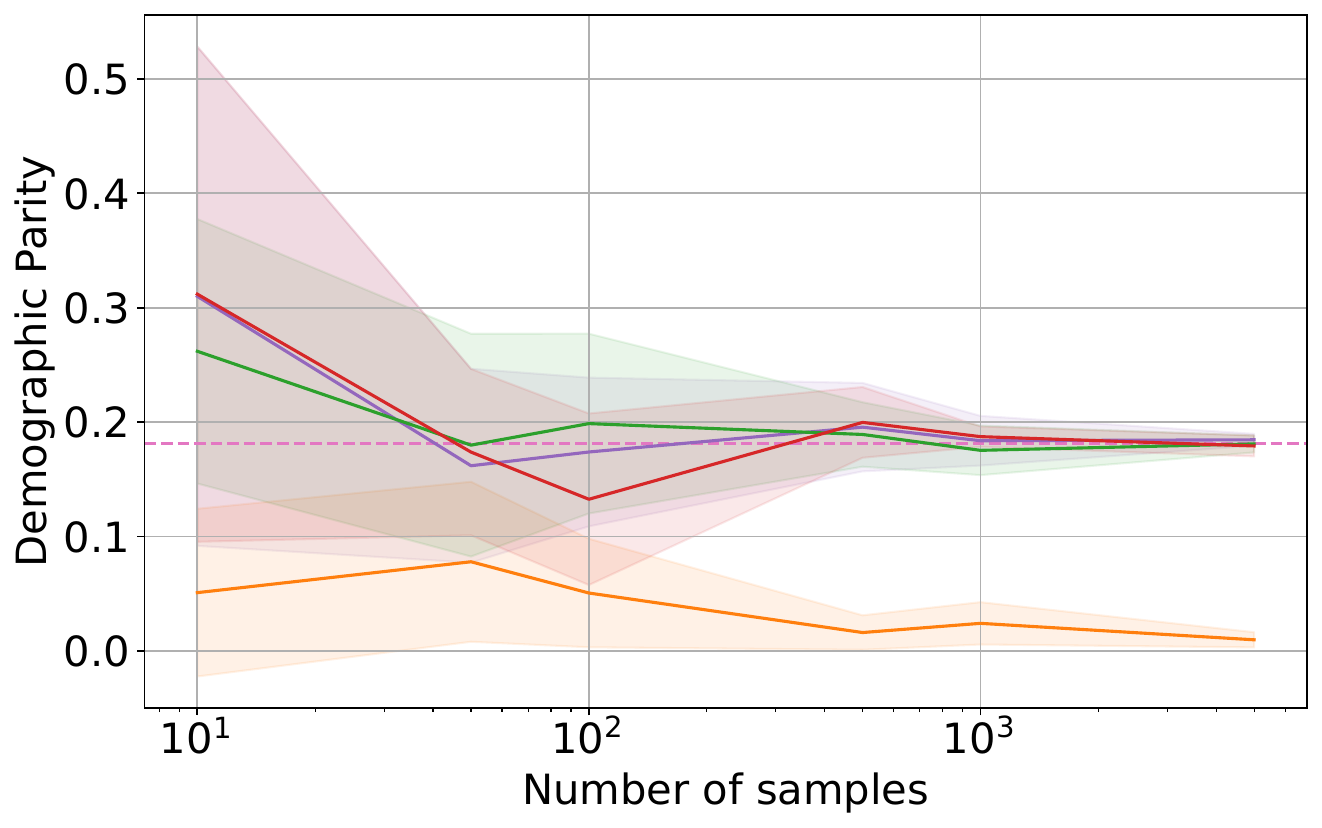}
  \caption{Adult}
\end{subfigure}%
\begin{subfigure}{.5\textwidth}
  \centering
  \includegraphics[width=\linewidth]{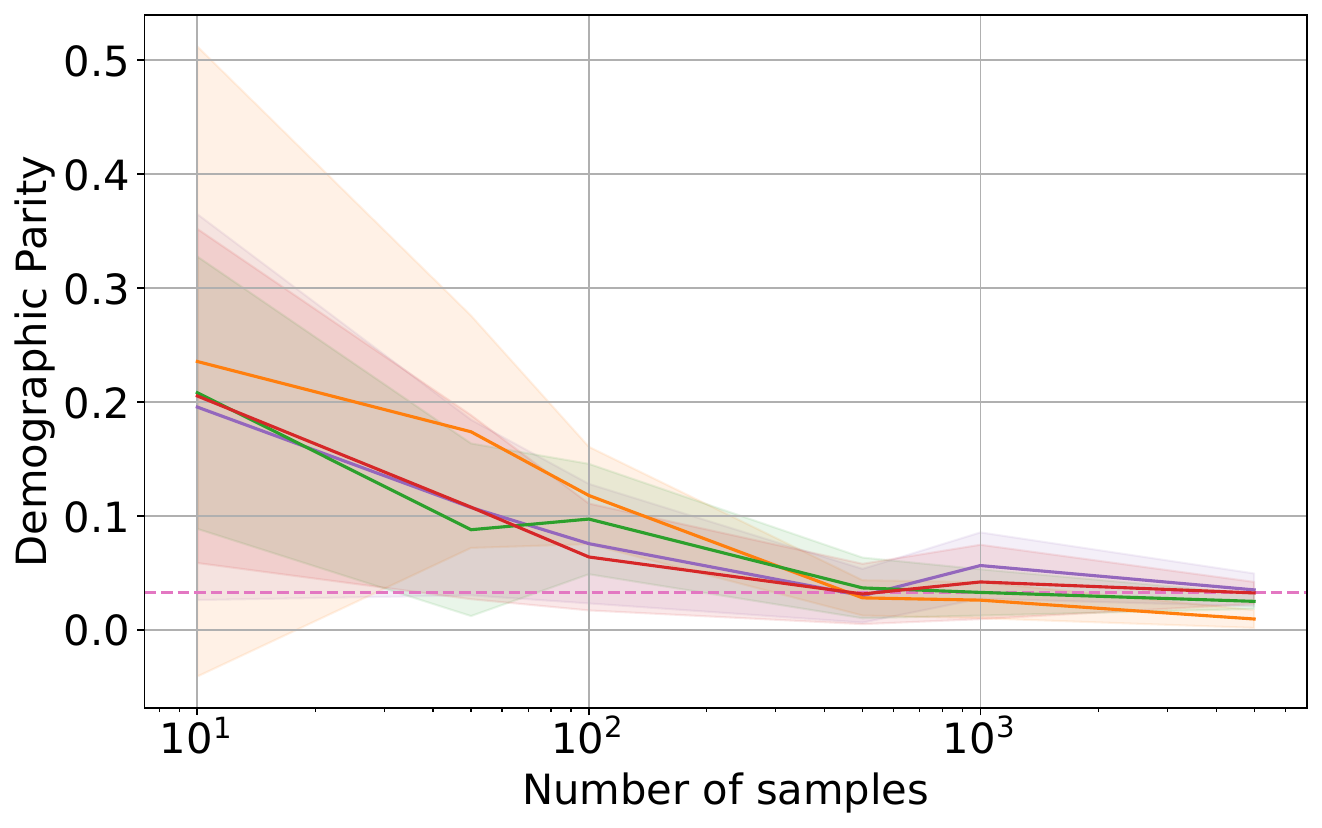}
  \caption{Folktables}
\end{subfigure}

\begin{subfigure}{\textwidth}
  \centering
  \includegraphics[width=.9\linewidth]{pics/legende_samples.pdf}
\end{subfigure}%

\caption{Demographic parity depending on the number of samples with XGB.}
\end{figure}

\begin{figure}
\centering
\begin{subfigure}{.5\textwidth}
  \centering
  \includegraphics[width=\linewidth]{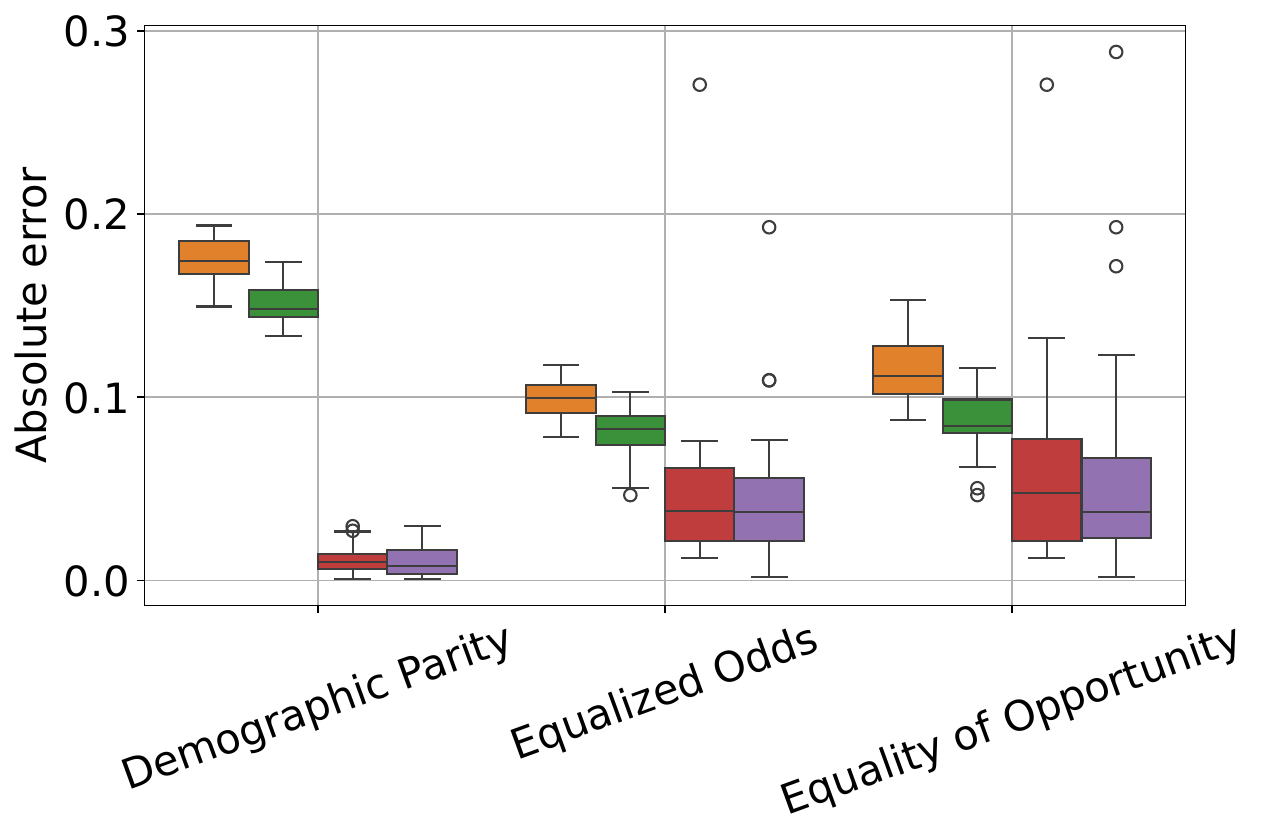}
  \caption{Adult}
\end{subfigure}%
\begin{subfigure}{.5\textwidth}
  \centering
  \includegraphics[width=\linewidth]{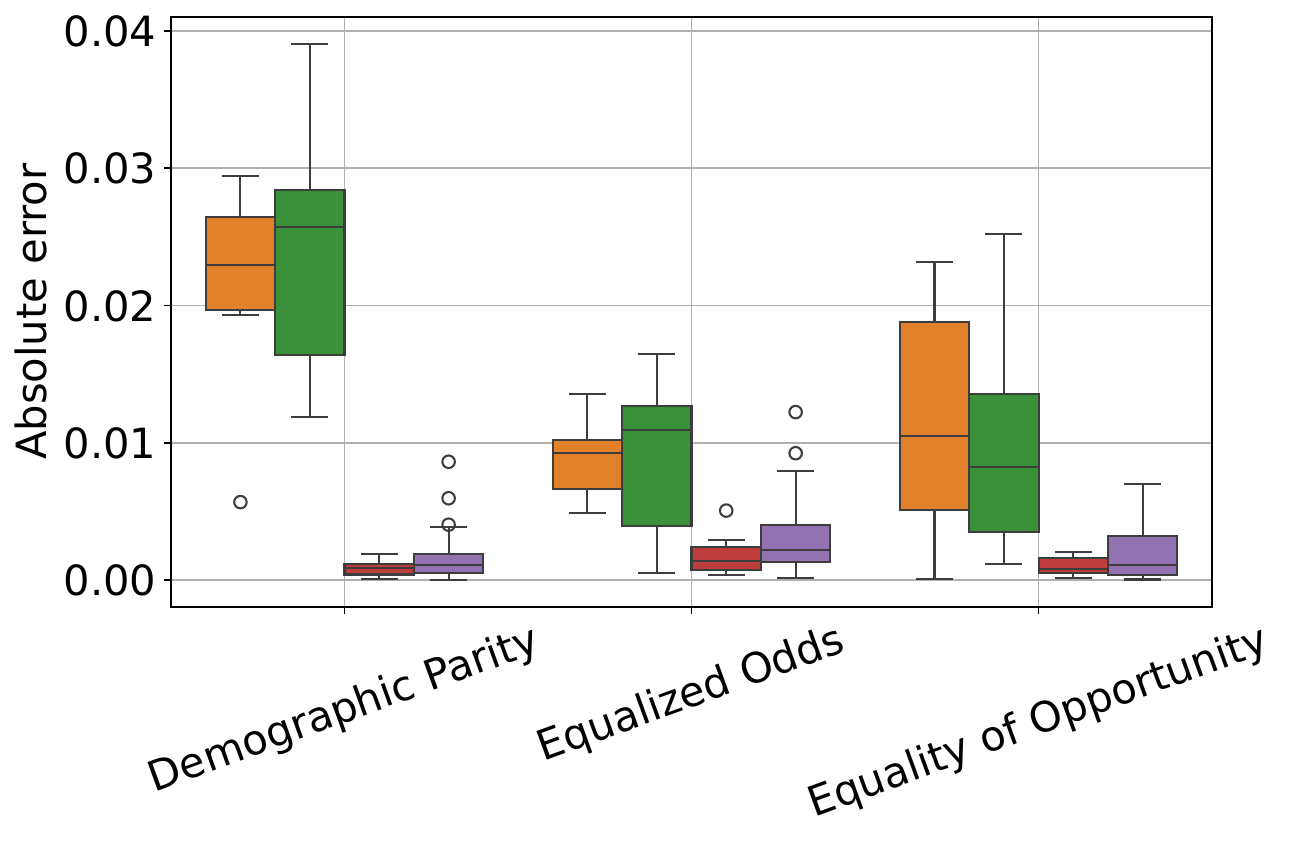}
  \caption{Folktables}
\end{subfigure}

\begin{subfigure}{\textwidth}
  \centering
  \includegraphics[width=.9\linewidth]{pics/legende_box.pdf}
\end{subfigure}%
\caption{Absolute difference compared to the reference for the three standard fairness metrics with $\epsilon = 1$, averaged on two models.}
\end{figure}

\end{document}